# Democratising Camera Trap AI: An Open-Source Model for Detecting UK Mammals

[1]Paul Fergus, [2,3]Philip Stephens, [2,3]Russell A. Hill, [4]Lee Oliver, [5]Katie Appleby, [6]Sarah Beatham, [1,7]Naomi Davies Walsh, [7]Stuart Nixon, [7]Naomi Matthews, [8]Chris Sutherland, [9]Kelly Hitchcock

[1]Liverpool John Moores University, Byrom Street, Liverpool, L3 3AF, UK.
[2]Durham University, Stockton Road, Durham, DH1 3LE.
[3]MammalWeb, c/o Durham Wildlife Trust, Houghton-le-Spring, DH4 6PU
[4]Game & Wildlife Conservation Trust, Fordingbridge SP6 1EF, UK.
[5]National Trust, Heelis, Kemble Drive, Swindon SN2 2NA.
[6]Animal and Plant Health Agency, Sand Hutton, York, Y04 1LZ.
[7]Chester Zoo, Caughall Road, Upton-by-Chester, Chester, CH2 1HU.
[8]University of St Andrews, College Gate, St Andrews, KY16 9AJ.
[9]Nottingham Trent University, Southwell, NG25 0QF.

Corresponding Author: p.fergus@ljmu.ac.uk.

***Abstract*— Camera traps have become a cornerstone of biodiversity monitoring, but the artificial intelligence that turns vast quantities of images into usable ecological data is often locked behind commercial platforms or trained on fauna that does not match that of the British Isles. In an attempt to remove barriers and increase uptake, we release an open-source object detection model for 31 classes, 28 common UK mammal and bird species, plus utility classes for humans, calibration poles, and vehicles, drawn from a curated dataset of 48,165 labelled instances assembled from multiple sites over a decade of operational deployment through Conservation AI and its successor, Trap Tracker. The model, a YOLO26x detector trained and tested on an 80/10/10 class-stratified split, achieves a mean Average Precision of 0.984 at Intersection over Union (IoU) of 0.5 (0.956 at IoU 0.5-0.95) on the held-out validation set, with precision 0.988 and recall 0.965. On an unseen held-out test split, mean per-species confidence ranged from 0.96 to 0.99 across the 31 classes, with a 0.17% false-negative rate concentrated in difficult night-time, distant, or occluded images. These metrics are from data from the same pool of sites and cameras as training, so performance at entirely new sites is left to future work. We release the trained weights in ONNX format under a non-commercial licence, with local desktop and real-time camera support, aimed explicitly at ecologists with no machine-learning experience. This release is a deliberate counterweight to the multiple paid for models that have developed over the last decade.**

## I. Introduction

Camera traps are highly effective for detecting wildlife, often outperforming methods such as line transects, live traps, and scat surveys [1], [2]. Their limitation has never been data collection but data processing: a single mid-sized UK deployment routinely produces tens of thousands of images per month, and continental aggregations run into the tens of millions [3], [4]. Small single-species studies can still be reviewed by hand, but at survey and landscape scale manual review collapses, and even citizen-science platforms, which have hand-classified images into the millions, cannot, in principle, keep pace with the rate at which cameras collect.

The consequence is not merely inconvenience. Imagery that is expensive to collect and ecologically valuable goes unprocessed; monitoring questions that the data could answer go unanswered; and the profound gains that automated analysis now makes possible are, for many organisations, simply not realised. Historically, organisations met this burden in one of two ways. Many kept annotation in-house, with small expert teams labelling hundreds of thousands or millions of images by hand; this preserved control over sensitive data and ensured consistent, expert labelling, but it is exactly this manual burden that made conventional camera-trap analysis so slow and so difficult to scale. Others turned to citizen-science platforms, Snapshot Serengeti being the canonical example, demonstrating that volunteer classifiers could process millions of images at high accuracy [5]. But citizen science is not available to everyone; sensitive species locations, data-governance constraints, and the absence of a public-facing platform keep many projects from using it, and even where it is available, volunteer effort is a finite resource whose accuracy varies substantially with species, habitat, and camera configuration [6].

Deep learning offered a third path, and it now dominates the field. The application began in earnest with Norouzzadeh et al.'s 2018 demonstration that convolutional neural networks could identify, count, and describe behaviours for 48 species across the 3.2-million-image Snapshot Serengeti dataset, with classification accuracy exceeding 93% on within-distribution test data, comparable to volunteer classifiers but operating at machine speed [7]. This work, and the parallel emergence of the North American Camera Trap Images (NACTI) dataset and the MLWIC2 model [8], [9], established the standard approach that has dominated the field since: collect a large labelled dataset, train a deep classifier or detector on it, report accuracy figures in the high 90s, release the model. Willi et al. demonstrated similar results combining deep learning with citizen-science labels [10]. Subsequent work refined the methodology: Norouzzadeh et al. added active learning to reduce the labels required for new datasets [11]; Schneider et al. and others moved from whole-image classification to object detection, providing localisation as well as identification [12]; and recent work has begun to address the long-tailed class distributions characteristic of operational camera trap data [13].

A persistent and consequential finding throughout this literature is that within-distribution performance numbers are not what conservation practitioners experience in the field. Beery et al.'s "Recognition in Terra Incognita" was the first careful demonstration that camera-trap models trained on a fixed set of camera locations generalise poorly to new locations [14]: a model trained on a set of camera stations learns to associate species with the backgrounds, vegetation, and lighting characteristic of those specific stations rather than with the animals themselves. Tabak et al. showed the same effect concretely, with a model trained on US camera-trap data dropping to 81.8% accuracy when tested on Canadian images of the same species [9]. Whytock et al.'s subsequent work in Gabon, explicitly designed for cross-site evaluation, held out 23,868 images from 227 independent camera stations, including species not represented in training, and reported 90% overall accuracy on this out-of-sample set, strong but still below the within-sample accuracies typically posted by camera-trap models [15]. More recent work reframes the problem entirely, arguing that the cross-site framing favoured by the ML community misses the question conservationists actually care about: maintaining reliable recognition at a fixed site as ecosystems and camera deployments drift over time [26]. The honest conclusion is that high accuracy numbers should always be read alongside their evaluation protocol, and that almost any model deployed to a new partner site, with new cameras and new vegetation, will see some performance degradation relative to its training-distribution benchmark.

When the Conservation AI project began working on automated camera-trap classification in the mid-2010s [16], the available tooling consisted of a handful of academic releases, no consumer-friendly software, and almost no commercial activity. At its peak, the Conservation AI platform processed approximately four million camera-trap images per week, free at the point of use for partner organisations [17]. By the time the platform wound down as a public service in 2025 and was rebuilt as Trap Tracker [18], this picture had shifted markedly, and it is worth distinguishing two things that are often conflated: the detection and classification models themselves, and the platforms that integrate them into end-to-end upload-to-analysis services.

On the model side, the operational landscape has consolidated around a small number of open tools, the two most prominent being MegaDetector [19] and SpeciesNet [20]. MegaDetector, developed at Microsoft AI for Earth by Beery, Morris and Yang and now maintained as part of the PyTorch Wildlife project, has become the de facto standard for the detection step in camera-trap pipelines [19]. Its deliberately species-agnostic design, returning only "animal", "person", or "vehicle", is its core strength: by not committing to a species taxonomy it generalises remarkably well across geographies and is used by 60+ organisations worldwide [21], with Vélez et al. reporting F1 values between 0.87 and 0.96 across diverse deployment contexts [28]. The trade-off is that it returns no species labels, so any practitioner needing species-level outputs must bolt on their own classifier. SpeciesNet, developed at Google as the species-level backbone of the Wildlife Insights platform, addresses this by training an EfficientNet V2 M classifier on millions of images covering more than 700 species [22]; its strength is breadth, but for a UK user that breadth is also its weakness, UK species are largely present in its class list, yet they sit within a global taxonomy of hundreds of species that do not occur here, the model is not tuned to the UK subset or to UK imaging conditions, and a large global class space introduces opportunities for confusion with non-UK congeners that can depress per-region precision and recall. The closest regional analogue, the European-focused DeepFaune classifier [23], is free, open, and runs locally, but its taxonomy mixes species-level and higher-order group classes, birds, for example, are grouped into a single class rather than resolved to species, which limits its usefulness for UK surveys that need species-level identification. The broader landscape includes the Wildlife Insights platform itself [22], which aggregated 4.5 million records at launch; MLWIC2 [9] and CameraTrapDetectoR [24], delivered as R packages; Camelot [25]; Mbaza AI [15], deployed in Gabon, designed for offline laptop use in low-bandwidth field conditions; AddaxAI (formerly EcoAssist), a free, open-source, offline wrapper around MegaDetector that now bundles European and sub-Saharan classifiers but no UK-specific model [26]; PyTorch Wildlife [27]; DeepForestVision for African tropical forests [28]; and a growing number of bespoke detectors built on the YOLO architecture family [29]. Vélez et al.'s 2023 review remains the most useful published comparison across ease of use, programming requirements, and species-classification capability [30]; its conclusion is that no single platform dominates, and that choice depends substantially on the user's resources and target species.

On the platform side, a commercial layer has grown up around camera-trap AI. Here it is important to be precise about what is and is not enclosed: the leading general-purpose models are free and open, and a capable practitioner can download and run them at no cost. What is increasingly commercialised is the layer that makes them usable, hosted processing, bulk-upload capacity, support, and the species-level, regionally-tuned classification that a general-purpose detector does not provide out of the box. Commercial platforms such as Animal Detect [31] package exactly this layer, building on the open models but charging on a credit or subscription basis, and consultancies offer camera-trap image processing as a billed service; even research-oriented platforms with free tiers, Wildlife Insights [22] and Agouti [32] among them, meter usage or charge for processing beyond a threshold. There are legitimate reasons for this, sustaining model development, providing user support, funding ongoing labelling. and we make no claim that commercial pipelines are inappropriate. The model we release sits deliberately on the open side of that line: open weights, runnable independently of any platform, free for non-commercial use.

Together these developments have produced a peculiar inversion. The technical ceiling has risen sharply, while practical access for small NGOs, university research groups, rewilding initiatives, citizen-science networks, and underfunded statutory agencies has narrowed. Globally the disparity is starker still: the regions facing the highest risk of mammalian defaunation are precisely those with the least technical expertise, capacity, and infrastructure [33]. A paywall on usable, species-level AI therefore falls hardest on exactly the under-resourced, high-biodiversity-risk regions least able to absorb it, particularly when the underlying training data have often come from the conservation community itself.

The case is especially acute in the UK, which despite being one of the most data-rich biodiversity-monitoring contexts in the world has scarce open, UK-tuned species-level tooling. Long-running citizen-science schemes coordinated by the People's Trust for Endangered Species [34], the Mammal Society [35], and the National Biodiversity Network [36] generate millions of records annually, and projects such as MammalWeb [37] have made substantial use of crowdsourced classification of UK camera trap imagery [38]. Yet the State of Nature 2023 report documents continuing decline, with 26% of UK terrestrial mammal species now at risk of local extinction and the country classified as one of the most nature-depleted in the world [39]. The implication is sharpened by policy: under the Environment Act 2021, the UK government carries a legally-binding duty to halt the decline in species abundance in England by 2030 and to increase it by 2042 [40]. Yet the official species-abundance indicator against which that target is judged is taxonomically lopsided, of the ~213 species in the relative-abundance measure, the great majority are birds and moths, with mammals represented by only around a dozen species, and those skewed toward bats, a hare, a dormouse and a water vole rather than the medium-to-large terrestrial mammals (deer, fox, badger, mustelids, wild boar) that camera traps are uniquely well-suited to survey [41], [42]. The country is thus committed by law to reversing biodiversity loss while measuring mammal abundance through an instrument that barely covers most of its mammals, making accessible, scalable, species-level camera-trap monitoring not merely convenient but a live route to closing a statutory monitoring gap.

The same pattern holds internationally and for a different driver of loss. Invasive non-native species are among the principal causes of global biodiversity decline, and Target 6 of the Kunming-Montreal Global Biodiversity Framework, agreed at COP15 in 2022, commits signatories, the UK among them, to reducing the rate of introduction and establishment of invasive alien species by at least 50% by 2030, alongside the control or eradication of those already established [43]. Meeting that target depends on knowing where invasive species are and in what numbers, yet the framework's headline indicator for invasives has been criticised for lacking any practical monitoring mechanism [44]. Several of the classes in the model released here, grey squirrel (*Sciurus carolinensis*), Reeves's muntjac (*Muntiacus reevesi*), and wild boar (*Sus scrofa*), are precisely the established non-native species whose distribution and abundance such targets require us to track, and which camera-trap monitoring is well-placed to deliver.

Recent UK-focused work confirms both the appetite and the gap. Carl et al. evaluated pre-trained European mammal detectors on German forest data with mixed results [45]; Sharpe et al. [46] and Hitchcock et al. [47] evaluated the Conservation AI UK Mammals model specifically, reporting strong precision for some species (fox, hedgehog >0.80) but recall issues that improved only after retraining, and concluding that accuracy remained below expert human performance without further refinement; and Davies Walsh et al.'s safari-park study, using a YOLOv10x detector on the Conservation AI desktop application, reported >0.90 accuracy across four species with strong correlation against human counts [48], with closely related work integrating YOLO detection and vision-language models to produce ecologically interpretable reports rather than raw labels [49]. Taken together, this establishes UK camera-trap AI as a live research area but one without a definitive open species-level model, direct evidence that even a UK-trained model has real headroom on UK data, and that the UK case warrants continued, UK-specific development.

This paper is a response to that gap. Trap Tracker has released, under a non-commercial Creative Commons licence, the trained weights and supporting tools for a UK-mammal-focused camera-trap detector, the first in a planned series of regional, species-specific open models, with a sub-Saharan African model in active preparation. The release is UK-specific by design: as the generalisation literature above makes clear, geographic and faunal alignment between training data and deployment context matters, and no existing open model is both tuned to UK fauna and packaged for a non-specialist. Equivalent gaps exist for every other major faunal region, and we intend to address them in turn.

The paper makes three substantive contributions. First, it provides ONNX weights for a YOLO26x detector covering 31 classes relevant to UK biodiversity, with a documented training procedure and full reporting of evaluation metrics. Second, it accompanies the weights with local desktop processing utilities and real-time camera-trap support aimed at ecologists rather than ML engineers, so that the model is usable by a conservation practitioner who has camera-trap images and a laptop but no budget for a commercial pipeline. Third, it frames this release as part of a broader case that publicly funded conservation, and the small NGOs, citizen-science networks, rewilding projects, and statutory agencies that depend on it, should not be pushed onto commercial platforms as their only option. We make no claim of methodological novelty: YOLO26x is off-the-shelf and the training is standard supervised learning at scale. The contribution is practical, a curated, UK-specific, openly-licensed detector built from operational partner data and returned to the community that produced it.

## II. Dataset

The dataset consists of 48,165 labelled animal instances drawn from camera trap images across UK sites, supplemented with additional images covering classes underrepresented in operational data. Throughout this paper, an *instance* refers to a single labelled object (one bounding box), not an image; a single camera-trap image may contain several instances (for example, multiple animals in one frame). The class taxonomy comprises 31 entries (see Table I):

- 28 animal classes — 18 mammal classes (including domestic and livestock species) and 10 bird classes
- 2 utility classes — Person and Calibration Pole, both important for calibrating and filtering camera trap output
- 1 reserved class — Car, defined in the taxonomy but not currently populated with labelled instances; the model will not predict this class in practice

Although the model is named for and dominated by mammals, its taxonomy is not restricted to them. It includes a small number of birds, domestic and livestock species, a Person class, and a Calibration Pole utility class. This reflects the operational reality the model was built for: camera-trap surveys do not photograph mammals in isolation, and a detector that ignores the people, livestock, and birds that share those frames is less useful, not more. Classes are included where a partner survey required them and labelled data existed, the Capercaillie and curlew classes are examples, rather than to satisfy a fixed taxonomic definition [50].

TABLE I
CLASSES INCLUDED IN THE DATASET AND LABEL COUNTS

| Common Name | Scientific Name | Labels |
|---|---|---|
| **European hedgehog** | *Erinaceus europaeus* | 2,131 |
| **Grey squirrel** | *Sciurus carolinensis* | 2,089 |
| **Red squirrel** | *Sciurus vulgaris* | 1,980 |
| **Roe deer** | *Capreolus capreolus* | 1,780 |
| **Red deer** | *Cervus elaphus* | 1,180 |
| **Red fox** | *Vulpes vulpes* | 2,391 |
| **European badger** | *Meles meles* | 1,801 |
| **Person** | *Homo sapiens* | 1,809 |
| **Domestic goat** | *Capra hircus* | 722 |
| **European rabbit** | *Oryctolagus cuniculus* | 1,810 |
| **Wood pigeon** | *Columba palumbus* | 1,625 |
| **Common pheasant** | *Phasianus colchicus* | 1,854 |
| **Domestic sheep** | *Ovis aries* | 1,232 |
| **House sparrow** | *Passer domesticus* | 1,410 |
| **Fallow deer** | *Dama dama* | 1,973 |
| **Domestic cattle** | *Bos taurus* | 1,353 |
| **Domestic cat** | *Felis catus* | 2,120 |
| **Pine marten** | *Martes martes* | 1,629 |
| **Domestic dog** | *Canis familiaris* | 1,909 |
| **Calibration Pole** | Calibration Pole | 151 |
| **Northern goshawk** | *Accipiter gentilis* | 1,598 |
| **Common buzzard** | *Buteo buteo* | 1,692 |
| **Capercaillie cock** | *Tetrao urogallus* | 2,005 |
| **Capercaillie hen** | *Tetrao urogallus* | 1,919 |
| **Eurasian curlew** | *Numenius arquata* | 1,451 |
| **Curlew chick** | *Numenius arquata* | 274 |
| **Car** | Car | — |
| **Domestic horse** | *Equus caballus* | 1,177 |
| **Carrion crow** | *Corvus corone* | 1,309 |
| **Wild boar** | *Sus scrofa* | 1,852 |
| **Reeves' muntjac** | *Muntiacus reevesi* | 1,939 |

## A. *Training, Validation and Test Sets*

The dataset is split using an 80/10/10 partition, i.e., 80% used for training (38,522 instances), 10% for validation (4,805 instances), and 10% for testing (4,838 instances). The three together comprise the full 48,165-instance dataset, which uses class-stratified sampling: each class is independently partitioned in the same 80/10/10 ratio (to the nearest 0.1%), so every class, including the smallest ones such as Curlew chick (219/27/28) and Calibration Pole (120/15/16), is meaningfully represented in all three splits. The validation split is used during training for model selection; the test split is held back entirely and used only for the post-training evaluation reported in Section V.

It should be noted that this constitutes a class-stratified hold-out derived from the same data pool as the training set, as opposed to a site-disjoint or temporally disjoint hold-out. The headline metrics below are therefore 'in-distribution' evaluations.

## B. *Class distribution and imbalance*

The number of labelled examples per class ranges from a low of 151 (Calibration Pole) and 274 (Curlew chick) and a high of 2,391 (Red fox) with most species having between 1,000 and 2,000 instances. The dataset was curated toward greater balance during assembly, and is therefore less skewed than typical operational camera trap data, where species such as grey squirrel (*Sciurus carolinensis*) or European rabbit (*Oryctolagus cuniculus*) can outnumber rarer species by orders of magnitude.

## C. *Image conditions*

The training set spans the conditions that any UK camera trap deployment will likely encounter,. Including representative daylight RGB, infrared night-time monochrome, motion blur, partial occlusion, multi-animal frames, and the manufacturer overlays and timestamps that ride along the bottom of nearly every camera trap photograph. Sample training mosaics are provided in Appendices at the end of the paper.

# III. Model and Training

We use YOLO26x, the extra-large variant of the YOLO26 object detection family, initialised from publicly released pretrained weights [51]. The decision to use YOLO26 rather than a more bespoke architecture is deliberate: object detection on camera-trap data is no longer an open research problem at the architecture level. What now determines whether a model is useful in the field is not the network design but the data it is trained on, the honesty with which its accuracy is reported, and the ease with which a non-specialist can deploy it – aspects we prioritise here. YOLO26 is a well-supported family with efficient inference on commodity GPUs, clean export to a portable, framework-independent format (ONNX) for easy deployment, and a path to lighter variants (YOLO26n, s, m, l) for users who want to trade accuracy for speed.

The model was trained on eight NVIDIA RTX A6000 GPUs with a global batch size of 256 at input resolution 640×640, mixed-precision (AMP), for 120 epochs, with cosine learning rate disabled and early stopping disabled (patience set high enough that full schedule completed). The full training run took 8 hours 8 minutes, approximately 244 seconds per epoch. Key hyperparameters included an initial learning rate of 0.01, final learning rate factor of 0.01, momentum of 0.937, weight decay of 0.0005, a 3 epoch warmup, mosaic augmentation closed at epoch 110, horizontal flip probability of 0.5, an HSV jitter (h=0.015, s=0.7, v=0.4), a random affine scale of ±0.5, translate of ±0.1, RandAugment enabled, and a random erasing of 0.4. Loss weights followed defaults (box=7.5, cls=0.5, dfl=1.5).

Training was performed with seed of 0 and deterministic equals true in the Ultralytics framework. We release the converted ONNX weights. We do not release the raw training images, as a substantial proportion are held under partner-specific data agreements that pre-date this open release.

# IV. Results

We evaluate the performance and operational behaviour of the 31-class UK camera-trap detection model, reporting headline validation metrics, training convergence, class-level confidence behaviour, qualitative examples, and an additional test-set evaluation. Together, these results characterise the model's behaviour across all classes and main remaining limitations.

### A. Headline Metrics

On the validation split, we report standard object-detection metrics (see Table II). Precision is the fraction of detections that are correct; recall is the fraction of animals actually present that are detected. The F1 score is the harmonic mean of precision and recall, and the operating point is set at the confidence threshold that maximises it. Metrics are reported at this F1-optimal threshold of 0.62; users running pipelines intolerant of false positives may prefer a higher threshold (0.75–0.85).

Mean Average Precision (mAP) is the field-standard summary of detector performance. For each class, Average Precision captures how well the model maintains both high precision and high recall as its confidence threshold is varied, rewarding models that find most animals while raising few false alarms, and mAP is the mean of this across all classes. A detection only counts as correct if its predicted box overlaps the true box by enough, measured as Intersection-over-Union (IoU). We report two variants: mAP@0.5 requires at least 50% overlap, a lenient test that essentially asks whether the animal was found and roughly located; mAP@0.5:0.95 averages performance over progressively stricter overlap thresholds (0.5 to 0.95) and so additionally rewards tight, accurate localisation. The second is the more demanding measure and is where continued training tends to show its effect.

TABLE II
HEADLINE VALIDATION METRICS FOR THE FINAL MODEL

| Metric | Value |
|---|---|
| Precision | 0.988 |
| Recall | 0.965 |
| mAP@0.5 | 0.984 |
| mAP@0.5:0.95 | 0.956 |
| Confidence threshold at peak F1 | 0.620 (F1 = 0.98) |

For context: an mAP@0.5 of 0.984 on a 31-class detection task with this much intra-class visual variability, diurnal vs nocturnal, RGB vs IR, juvenile vs adult, partial vs full body, is consistent with the strongest figures reported for UK-focused systems, though direct comparison is complicated by differing evaluation protocols. This comparison is made with a clear caveat: these are in-distribution numbers, as are most of the published comparisons.

### B. How the Model Learned During Training

The training and validation curves (Fig. 1) show performance rising steeply in the first few epochs and then settling at a stable level, with validation losses continuing to fall and then flattening rather than rising again. The absence of any late upturn in validation loss, the signature of a model that has begun memorising its training data rather than learning generalisable features (overfitting), indicates the model converged cleanly within its training distribution. Most of the headline performance is achieved by approximately epoch 50, with mAP@0.5 increasing from 0.885 at epoch 1 to 0.981 by epoch 50. Continued training to epoch 120 produces only a small additional gain in mAP@0.5, reaching 0.984, but a more meaningful improvement in the stricter mAP@0.5:0.95 metric, which rises from 0.933 at epoch 50 to 0.956 at epoch 120. This suggests that the later epochs primarily refine localisation quality rather than producing large gains in coarse detection performance. Training and validation losses for the three components of the detector's loss function, bounding-box regression (box), classification, and distribution focal loss (DFL), which refines box-edge localisation, decline steadily after the initial warmup period.

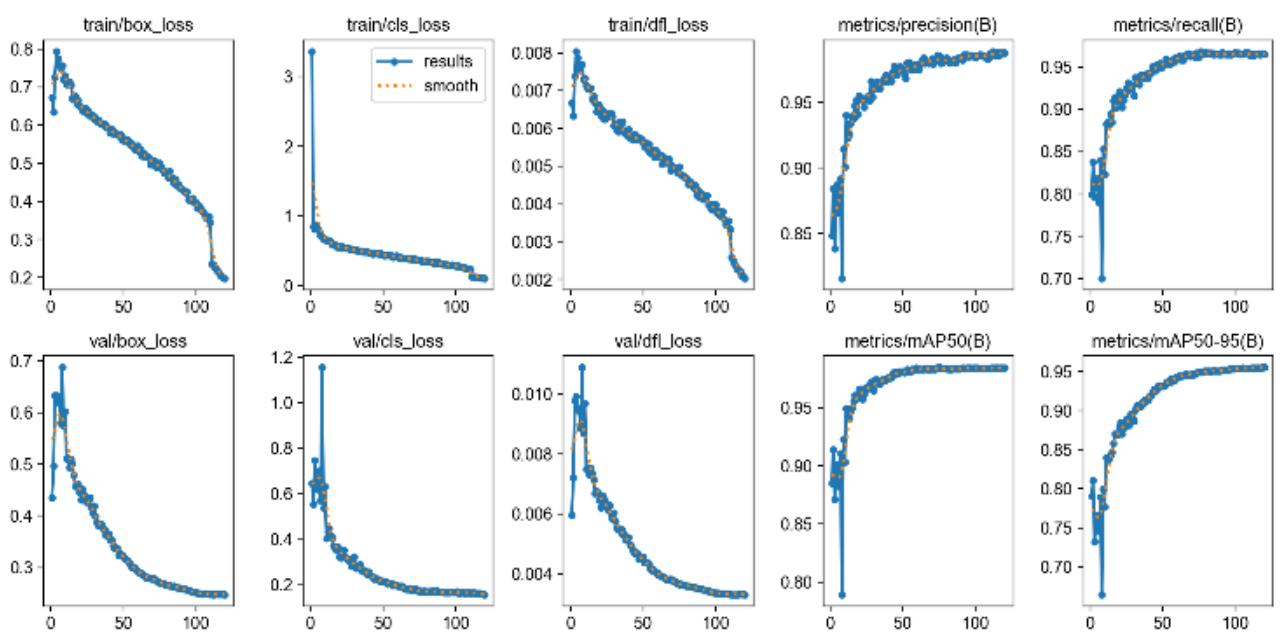

Fig. 1. Training dynamics over 120 epochs. The x-axis of every panel is the training epoch. Top row: training losses for the three components of the detector's objective, bounding-box regression (box_loss), species classification (cls_loss), and distribution focal loss (dfl_loss), which refines box-edge localisation. Bottom-left three panels: the same three losses computed on the held-out validation split (val/box_loss, val/cls_loss, val/dfl_loss). Right-hand panels (top and bottom): validation-set detection metrics, precision, recall, and mean Average Precision at IoU 0.5 (mAP@0.5) and averaged over IoU 0.5–0.95 (mAP@0.5:0.95); the "(B)" suffix denotes bounding-box metrics as reported by the Ultralytics framework. Blue points are per-epoch values; the dotted line is a smoothed trend. All loss curves decline and flatten without a late upturn in validation loss, and all metrics rise steeply before plateauing, indicating clean convergence with no sign of overfitting within the training distribution.

### C. Per-class behaviour across full dataset

Confidence here is the model's own reported certainty in its top detection, not a measure of correctness against ground truth: a detector can be confidently wrong. What the per-class confidence distributions show is whether the model produces decisive, stable predictions across all 31 classes, including the rare ones, rather than returning consistently low or uncertain scores for some species or failing on particular classes altogether. Verified accuracy against hand-labelled ground truth is a separate question, addressed for related models in our prior work [46], [47] and identified as the priority benchmark for this release in Section IX.

To provide a class-by-class view that the headline validation metric does not, the trained model was run across the entire dataset (all 48,165 labelled instances), and best-detection confidence was recorded per image. Across these images, spanning all 31 classes, mean per-species confidence was 0.95–0.99, with median confidence of 0.97–0.99. Only 13 detections (0.027%) fell below confidence 0.4, that is, the model is almost never uncertain on data drawn from its training distribution. Mean confidence remains above 0.95 even for the smallest classes (Curlew chick: n=274, mean=0.969; Calibration Pole: n=151, mean=0.954).

### D. Qualitative results

Appendix A shows the training batches with their ground-truth labels, and Appendix B shows side-by-side ground-truth labels and model predictions on validation batches that were held out from weight training. The model produces tight, well-calibrated bounding boxes across diverse conditions: diurnal close-ups, infrared night frames, partially occluded animals, and multi-animal frames. Appendix C provides some detections using the model and real-time camera application.

## E. Test set evaluation

The previous metrics are computed on the validation split, which, although the model was never trained on it, was used during training for model selection (i.e., checkpoint choice). To provide an evaluation cleanly separated from training-time decisions, we additionally evaluated the model using a test split (10% of the dataset, 4,838 instances). The test split provides a check:

- of whether the model's confidence behaviour is stable on data unseen during training or validation
- that every class, including the very small ones, is usable, not just the well-represented ones
- the model is not overfitting to the validation split via inadvertent selection pressure

On the test data, the model produced confident detections in 4,812 of the 4,820 images, spanning all 31 classes (the difference between the 4,838 ground-truth instances and the per-image detection counts reflects multi-instance images, where one image contains several labelled objects). Mean per-species confidence ranged from 0.959 to 0.988 (see Table III).

The remaining 8 images (0.17%) fall into a residual 'Don't Know' category, i.e., images that contain a ground-truth object the model failed to detect at the operating threshold. These are not cases of the model guessing wrongly, but of it producing no confident prediction at all, true misses (false negatives), concentrated in the hardest imaging conditions: night-time long-distance shots, heavy occlusion, and severe motion blur. A single further detection, a domestic cat (*Felis catus*) called at 0.338 confidence, was the only confident-threshold species prediction to fall below 0.4. Taken together, only 9 of the 4,820 images (0.19%) produced a detection below 0.4 confidence.

The pattern is worth stating clearly, because it affects how the model should be used. In the hardest cases, distant or heavily occluded animals, or poor night-time exposures, the model tends to produce no detection at all rather than a confident but incorrect species label. These are true misses (false negatives): the object is present but absent from the model's output, not assigned to a low-confidence or 'unknown' category. For a conservation workflow this is the safer failure direction, a missed detection can be caught by manual review of empty frames, whereas a confident misclassification can silently corrupt a dataset, but it does mean the model will under-count in difficult conditions, consistent with the recall limitations identified in independent evaluations [46], [47].

The meaningful result is consistency across the class taxonomy. Every class, including those with very small representation in the test split, produced mean confidence above 0.95. This is evidence that the model has learned all 31 classes robustly within its training distribution and that no class is being carried by overfitting to a handful of validation examples.

TABLE III
PER-SPECIES CONFIDENCE SUMMARY ON THE TEST SET ORDERED BY SAMPLE SIZE

| Species | n | Mean | Median | Min |
|---|---|---|---|---|
| ***Vulpes vulpes*** **(red fox)** | 236 | 0.972 | 0.980 | 0.712 |
| ***Meles meles*** **(European badger)** | 211 | 0.971 | 0.979 | 0.674 |
| ***Erinaceus europaeus*** **(European hedgehog)** | 209 | 0.975 | 0.982 | 0.558 |
| ***Sciurus carolinensis*** **(grey squirrel)** | 208 | 0.966 | 0.974 | 0.555 |
| ***Felis catus*** **(domestic cat)** | 208 | 0.968 | 0.980 | 0.338 |
| ***Dama dama*** **(fallow deer)** | 208 | 0.973 | 0.983 | 0.613 |
| ***Muntiacus reevesi*** **(Reeves's muntjac)** | 202 | 0.977 | 0.983 | 0.773 |
| ***Sciurus vulgaris*** **(red squirrel)** | 193 | 0.968 | 0.974 | 0.633 |
| ***Capreolus capreolus*** **(roe deer)** | 192 | 0.973 | 0.980 | 0.688 |
| ***Sus scrofa*** **(wild boar)** | 191 | 0.977 | 0.981 | 0.821 |
| **Tetrao urogallus, male (*Capercaillie*)** | 188 | 0.978 | 0.984 | 0.518 |
| ***Homo sapiens*** **(person)** | 180 | 0.982 | 0.986 | 0.854 |
| ***Phasianus colchicus*** **(common pheasant)** | 180 | 0.968 | 0.976 | 0.694 |
| ***Tetrao urogallus*, female (Capercaillie)** | 179 | 0.966 | 0.969 | 0.834 |
| ***Columba palumbus*** **(wood pigeon)** | 174 | 0.973 | 0.985 | 0.414 |
| ***Accipiter gentilis*** **(northern goshawk)** | 164 | 0.973 | 0.980 | 0.543 |
| ***Buteo buteo*** **(common buzzard)** | 161 | 0.980 | 0.982 | 0.840 |
| ***Canis familiaris*** **(domestic dog)** | 160 | 0.975 | 0.981 | 0.648 |
| ***Martes martes*** **(pine marten)** | 160 | 0.966 | 0.973 | 0.607 |
| ***Bos taurus*** **(domestic cattle)** | 147 | 0.982 | 0.987 | 0.748 |
| ***Oryctolagus cuniculus*** **(European rabbit)** | 143 | 0.978 | 0.980 | 0.937 |
| ***Numenius arquata*** **(Eurasian curlew)** | 141 | 0.988 | 0.990 | 0.960 |
| ***Passer domesticus*** **(house sparrow)** | 141 | 0.972 | 0.980 | 0.723 |
| ***Corvus corone*** **(carrion crow)** | 130 | 0.968 | 0.971 | 0.909 |
| ***Equus caballus*** **(domestic horse)** | 127 | 0.978 | 0.986 | 0.759 |
| ***Ovis aries*** **(domestic sheep)** | 125 | 0.982 | 0.985 | 0.907 |
| ***Cervus elaphus*** **(red deer)** | 123 | 0.973 | 0.981 | 0.880 |
| ***Capra hircus*** **(domestic goat)** | 80 | 0.969 | 0.980 | 0.446 |
| ***Numenius arquata*, juvenile (Curlew chick)** | 33 | 0.968 | 0.969 | 0.909 |
| **CalibrationPole (utility)** | 18 | 0.959 | 0.965 | 0.915 |

## V. Limitations and Caveats

The metrics reported are genuinely strong by standard performance metrics. However, a release of this kind only merits the trust of the conservation community to the extent that the work is transparent about both its contributions and its limitations. Against this backdrop there are several issues to consider.

The first concerns the nature of the evaluation. All metrics in this paper, validation and test alike, are computed on data drawn from the same overall pool as training, partitioned by class-stratified hold-out. This is appropriate evidence that the model has learned the 31 classes robustly and is not overfitting to a particular subset of images, but it does not, on its own, demonstrate that performance transfers to a new partner site, a new camera model, or a new seasonal context. The literature reviewed in this paper has been consistent on this point for nearly a decade: camera-trap classification/object detection models tend to learn the cameras as much as the animals, and headline accuracy figures from in-distribution evaluation routinely drop on first deployment to a novel context [8], [14], [15], [52]. Based on our group's prior operational experience processing data at scale, and on the consistency of confidence behaviour we observe on the test split here, we would expect

mAP@0.5 to remain comfortably above 0.9 in most operational UK conditions, but this paper does not prove that, and we identify a labelled cross-site benchmark as the most important piece of work moving forward. Related to this, our test evaluation reports confidence distributions and review-flagging rates rather than precision and recall against manually-labelled ground truth. We have produced fully labelled evaluations of this kind in prior work [53–55], and the same methodology applies here; the choice in this paper was to prioritise the open release and behavioural verification, with hand-labelled benchmarking to follow.

A second set of caveats concerns the dataset itself. The Curlew chick and Calibration Pole classes are represented by relatively few training instances (274 and 151 respectively), reflecting their operational rarity rather than any neglect. Encouragingly, both classes are detected with high confidence on the test set (mean confidence 0.968 and 0.959 respectively), suggesting the model has genuinely learned them rather than memorised a few examples; but the absolute number of unseen test instances is small (33 and 18), and practitioners deploying in contexts where these classes are critical should expect to validate locally. Similar caution applies to the Capercaillie classes, which we have split by sex (cock and hen) rather than treated as a single class. The split is useful for grouse monitoring, where sex ratios are an ecologically meaningful signal [50], but it does mean the model is performing a finer-grained classification task on these classes than on, for example, the deer, and reviewers should not read the Capercaillie confidence figures as if they were directly comparable to those for less morphologically variable species. The most ecologically predictable error is occasional confusion between roe deer (*Capreolus capreolus*) and red deer (*Cervus elaphus*), which share silhouette and pelage tones and are most easily confused at distance and under infrared illumination; practitioners running UK deer surveys should be alert to this. More broadly, the model is mammal-focused with a curated set of important bird and utility classes; it is not, and should not be used as, a general UK avifauna detector.

Two further caveats concern deployment context. The training set contains infrared imagery from several manufacturer sensors, but the IR illumination spectra and sensor characteristics of camera traps continue to evolve, particularly in newer thermal-IR units, and cameras whose IR profile sits outside our training distribution will likely require fine-tuning rather than zero-shot deployment. We discuss this in the future-work section as a candidate for a planned mechanism that would let partner organisations contribute new labelled data to improve future versions of the model. Geographically, "UK" in our title is shorthand for mainland Britain: the species list reflects what is realistically encountered on the partner sites that contributed training data, and we expect reasonable transfer to comparable Northwest European faunas, but we have not benchmarked this and would not claim it without doing so.

Taken together, these caveats describe the boundary of what the present release does. Inside that boundary the model is strong across all classes, with consistent confidence behaviour on previously unseen data and a false-negative rate of 0.17% in the regime tested. Outside of it, for novel sites, novel cameras, novel fauna, or applications where labelled accuracy is required for downstream statistical inference, the model is a starting point rather than a finished tool, and we continue to fine-tune different versions of the model accordingly.

## VI. Release: Weights, Tools, Tutorials

The release accompanying this paper consists of the trained model weights, the configuration needed to reproduce or interpret them, and the practical tooling we think a conservation practitioner needs to use the model on their own data.

The weights themselves are released in ONNX format under a Creative Commons CC BY-NC 4.0 licence. The choice of ONNX rather than a framework-specific format (PyTorch .pt, TensorFlow SavedModel) is deliberate: ONNX is well-supported across inference runtimes including ONNX Runtime, OpenCV's DNN module, and Microsoft DirectML, runs on CPU and GPU, and can be deployed on Windows, macOS, and Linux without pulling in the full training framework. The FP16 variant trades a small amount of precision for substantially reduced memory footprint and faster inference on GPU hardware, which matters for the practitioner running through tens of thousands of images on a laptop. The CC BY-NC licence is similarly deliberate. We discuss the reasoning behind that choice below, but the headline is straightforward, Trap Tracker remains the commercial vehicle for this lineage of work, and a non-commercial licence on the open weights protects the project's ability to continue funding development of the next models in the series.

Alongside the weights we release the class taxonomy specification (data.yaml), which fully defines the 31-class label space and the integer-to-species mapping used by the model, and the training configuration (args.yaml), which captures the exact hyperparameters used to produce the checkpoint: learning rate schedule, augmentation policy, loss weights, optimiser, batch size, image resolution, and the full set of Ultralytics framework settings. Together these allow a researcher to reproduce the training procedure, audit the model's behavioural assumptions, or use the configuration as a starting point for fine-tuning on their own data. Partners interested in collaborating can contact the authors directly.

The Python tooling is what most users will interact with day-to-day. We provide inference scripts for single-image and batch-folder use, with defaults tuned for camera-trap workflows rather than generic computer vision: sensible confidence thresholds for the operating point identified in the paper, appropriate handling of multi-animal frames, support for the directory structures that camera trap surveys typically produce, and clean CSV export of detections for downstream ecological analysis. Two interaction modes are supported. The first is a local desktop application that allows a user to point at a folder of images, run inference, and inspect the results visually; the second is a real-time camera mode that ingests imagery as it arrives from 3/4G-connected camera traps, mirroring the operational mode established in our prior work [55]. Both modes are wrapped in tutorials aimed explicitly at ecologists with no machine-learning background. The tutorials assume basic Python literacy but nothing more, and walk through installation, running inference on a folder of images, interpreting the confidence scores, and exporting results to CSV in a form that integrates cleanly with the statistical workflows ecologists use.

The intended audience for this work is primarily conservation practitioners, ecologists, and wildlife managers,

and the documentation has been written with these users in mind. Beyond that primary audience, three other groups are explicitly in scope. Academic ecology research groups will find here a strong baseline to fine-tune from for their own studies, with reproducible training configuration. Academic ecology research groups, citizen-science projects, and conservation NGOs alike can fine-tune from a documented baseline or run the model as a first-pass filter ahead of human review. And statutory agencies will find an inspectable model whose behaviour, training data, and limitations are documented and auditable, rather than a black box delivered as a SaaS subscription. The model can serve all audiences from a single release.

## VII. Discussion

The word "democratised" gets used loosely in AI press, often to mean little more than that something has been made available at zero financial cost. We mean something more specific in this paper. A conservation organisation with limited financial resources, no machine-learning staff, and no GPU should be able to run this model on a CPU laptop over a folder of camera-trap images, get species-level detections out as CSV, and feed those detections into whatever ecological analysis they were going to do. That is the standard against which the release should be judged. The weights are open. The licence is permissive for non-commercial use. Inference works entirely offline, on commodity hardware, with no network connection or vendor account required. The tutorials are written for ecologists rather than for machine-learning engineers. There is no platform tier above which features become paid, no per-image charge, no usage data is collected, and no quota imposed on how much imagery a partner can process. None of these individual properties is unique to this release, but in combination they describe what we think open conservation AI ought to look like.

This paper is not, and we do not present it as, a contribution to computer vision research. The novelty has nothing to do with the architecture, the optimisation, or the choice of detector; those are off-the-shelf, and a reader from the ML community will find nothing surprising in the technical sections. The work that matters, and the work that made this release possible, was the slow accumulation of well-labelled, multi-site UK camera-trap data over the better part of a decade, and the design of a class taxonomy that maps onto what UK conservation surveys need. Neither of those is the kind of thing that produces a research-paper headline, and both are exactly the kind of thing that gets held back when teams are reasoning about competitive advantage.

The current release is a function of where the field has moved over the last decade. When the Conservation AI project began, the interesting question was whether deep learning could process camera trap data accurately enough to be useful. By the time the public-facing platform wound down and migrated, the interesting question had become which vendor would charge what, and on what terms, and increasingly, for the UK practitioner wanting species-level results, the most usable options came with a price attached. This is not because capable models ceased to be free, MegaDetector, SpeciesNet and similar tools remain open, but because none of the free options was tuned to UK fauna and packaged for a non-specialist, leaving a paid service as the path of least resistance. We see this as a natural professionalisation of a once-immature field, and we are not arguing against commercial conservation AI as such. What we do think, and what this paper is arguing, is that the high-performance baseline should be freely available to anyone doing non-commercial conservation work, and we do not currently see another route to making that happen other than releasing it ourselves.

We would like this release to be the first of many such contributions rather than an exception. The argument of this paper is also not specific to the UK. It applies wherever usable, species-level AI for conservation is drifting behind paywalls, and the remedy it points to, releasing strong, region-specific weights openly, is one that any group holding such weights can adopt. We have already begun work on a sub-Saharan African camera-trap model under the same principles, building on the labelled evaluation framework we established with our 12-species Welgevonden deployment [49]; further regional models are in scope for future work. We encourage other groups in the camera-trap AI community who hold strong region-specific weights to consider equivalent open releases. The cost of doing so is small. The cumulative benefit to the conservation organisations, citizen-science networks, and statutory agencies that depend on accurate, accessible biodiversity monitoring is large, accruing fastest in the regions where it is most needed.

## VIII. Future Work

The most important immediate piece of work is a labelled out-of-distribution benchmark. The natural form is a hand-labelled test set drawn from partner sites, cameras, and seasons that are not represented anywhere in our training pool, scored with the same precision, recall, and per-species F1 methodology used in our previous papers. This is more useful than any architectural improvement we could make to the model itself, because it tells a practitioner what to expect when deploying on their own data rather than on ours, and we regard it as the principal weakness of the current release that needs closing.

Beyond that, several lines of work are already in scope. We intend to release smaller ONNX variants of the model (YOLO26n, s, and m) so that practitioners can choose an inference-speed-versus-accuracy trade-off appropriate to their hardware. Full YOLO26x is excellent on a GPU but heavier than necessary for many laptop and edge deployments. We will also publish documented fine-tuning recipes for partners who want to add region- or site-specific classes; the present model covers 31 classes broadly relevant across UK conservation contexts, but particular sites will want to add (for example) specific livestock breeds, local bird species, or vehicle types relevant to their monitoring questions, and the workflow for doing that should not require deep machine-learning expertise. A separate companion model focused on UK avifauna is also planned, since the current bird coverage in this release is restricted to a handful of important species rather than the full UK avian community.

A more ambitious direction we are actively exploring is a federated improvement mechanism, in which partners can contribute newly labelled data back to a successor model with appropriate governance over data ownership, labelling standards, and credit. The technical infrastructure for this is relatively straightforward; the greater challenge is the social and institutional design of it, which we expect to take longer than the technical development. We would rather prioritise getting this right over delivering it quickly

Underpinning all of the above is a commitment we want to make: the present release is not a one-off snapshot. We intend to continue releasing progressively improved versions of this UK mammals model as additional partner data accumulates, as labelling gaps in the long-tail classes are filled, and as we incorporate lessons from the out-of-distribution benchmark above. Each successor version will be released under the same non-commercial licence, with the same level of documentation and the same reproducible training configuration. Practitioners building on the present weights should expect a stable upgrade path rather than a moving target. Beyond the UK, the regional-release programme outlined in the paper is already underway, starting with a sub-Saharan African model and extending to further regions thereafter.

## ACKNOWLEDGMENT

The model released in this paper rests on data, expertise, and goodwill contributed by a wide range of UK partner organisations and individual collaborators. The acknowledgements below are organised by partnership area; we apologise in advance for any unintended omissions.

Liverpool John Moores University. Conservation AI was founded at LJMU and the present work continues in the same institutional home. We thank the team of co-developers and collaborators who have shaped the project across multiple papers, in particular Carl Chalmers, Prof. Serge Wich, and Prof. Steven Longmore, with whom much of the underlying Conservation AI infrastructure was developed. We also thank Naomi Davies Walsh and other LJMU research staff/students whose contributions to data labelling (particularly Rachel Chalmers), model evaluation, and field deployment have been essential.

Animal and Plant Health Agency (APHA). We thank APHA for the substantial contribution of camera-trap data over many years and for extensive independent evaluation of the UK Mammals model in operational wildlife-monitoring contexts, which has directly informed its development and refinement.

The Game & Wildlife Conservation Trust (GWCT) in Wales. The curlew data underpinning a substantial portion of the present model, and the operational deployment infrastructure for ground-nesting bird detection, was developed jointly with GWCT. We thank James Warrington, Julieanne Quinlan, and Katie Appleby for years of fieldwork, labelling, and partnership across the 11 Welsh nesting sites that informed the curlew components of this.

Durham University and MammalWeb. We acknowledge the broader collaboration with Durham University and the MammalWeb citizen-science platform, of which the National Hedgehog Monitoring Programme is the largest single project, and in particular Dr C. R. Sharpe, H. M. Chappell, Dr S. E. Green, K. Holden, and colleagues, whose independent evaluation work on earlier Conservation AI models has directly informed the present release. We are grateful for awards from the EPSRC IAA, ESRC IAA, and the Durham University Research Impact Fund.

Nottingham Trent University. We thank Dr Simon Tollington, Dr Richard Yarnell, Dr Leah Williams, and Dr Kat Hamill at NTU for independent evaluation of the Conservation AI UK Mammals model and for the semi-automated workflow analysis that has shaped our thinking on practitioner-facing deployment.

Chester Zoo and Knowsley Safari. We thank Chester Zoo for partnership across multiple Conservation AI projects involving ex-situ and in-situ monitoring, and Knowsley Safari for hosting field deployments of the YOLOv10 detector that informed the operational pipeline used in this release.

University of Aberdeen and University of St Andrews. We thank Prof. Xavier Lambin (Aberdeen), Dr Jack Bamber, Amber Cowans, and Oliver Hartley (St Andrews) for ecological supervision and methodological input on multi-site UK camera-trap deployments.

Technical partners. We thank NVIDIA for sustained support of the underlying compute infrastructure and for university partnership through the NVIDIA DLI programme, and Reolink for hardware partnership on the 3/4G camera platform that has underpinned several real-time deployments. We would also like to thank Vodafone for the generous SIM card donations over the years which has helped significantly in our UK studies.

Funders and supporting bodies. We acknowledge funding and support over the years from the UK Research and Innovation councils, NERC (including the IAPETUS DTP), ESRC IAA, EPSRC IAA, the Science and Technology Facilities Council (STFC), together with the Department for Environment, Food and Rural Affairs (Defra) and Natural England. We also acknowledge the network of UK wildlife trusts, statutory conservation agencies, rewilding initiatives, and individual landowners and farmers whose camera trap deployments have contributed to the curated dataset over the years. Many of these contributions came under partner-specific data agreements that do not permit individual attribution; we are no less grateful for them.

To anyone who has contributed time, data, expertise, or hardware to the Conservation AI/Trap Tracker programme and is not named explicitly above: thank you. This release would not exist without you.

## DATA AND CODE AVAILABILITY

Weights, configuration files, inference scripts, and tutorials are available at: *https://www.traptracker.co.uk*. Raw training images are not released.

**APPENDIX A:** SAMPLE TRAINING MOSAICS - SHOWING CONDITIONS THAT ANY UK CAMERA TRAP DEPLOYMENT WILL ENCOUNTER: DAYLIGHT RGB, INFRARED NIGHT-TIME MONOCHROME, MOTION BLUR, PARTIAL OCCLUSION, MULTI-ANIMAL FRAMES, AND THE MANUFACTURER OVERLAYS AND TIMESTAMPS

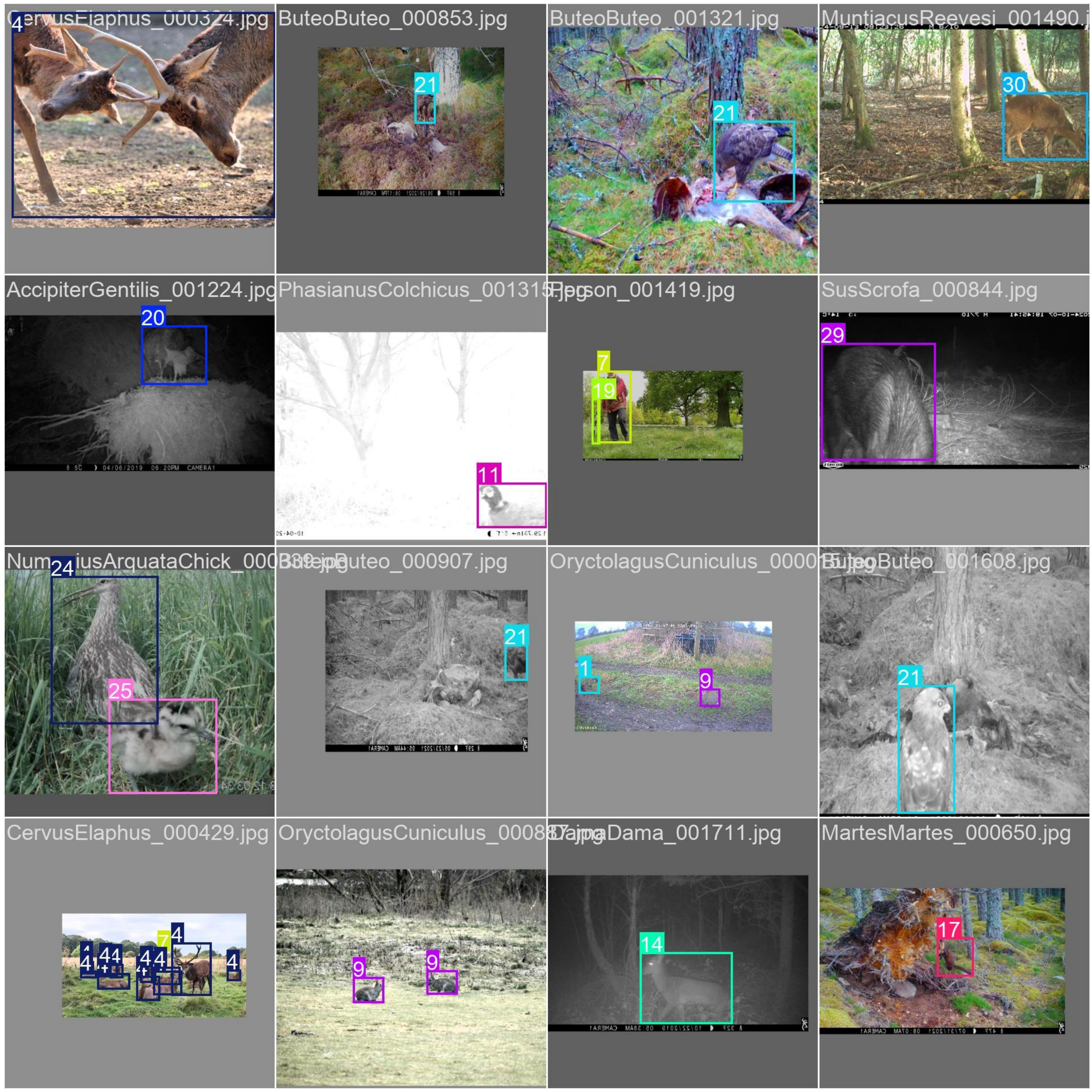

## **APPENDIX B:** SIDE-BY-SIDE GROUND-TRUTH LABELS AND MODEL PREDICTIONS ON VALIDATION BATCHES THE MODEL HAS NOT SEEN DURING TRAINING

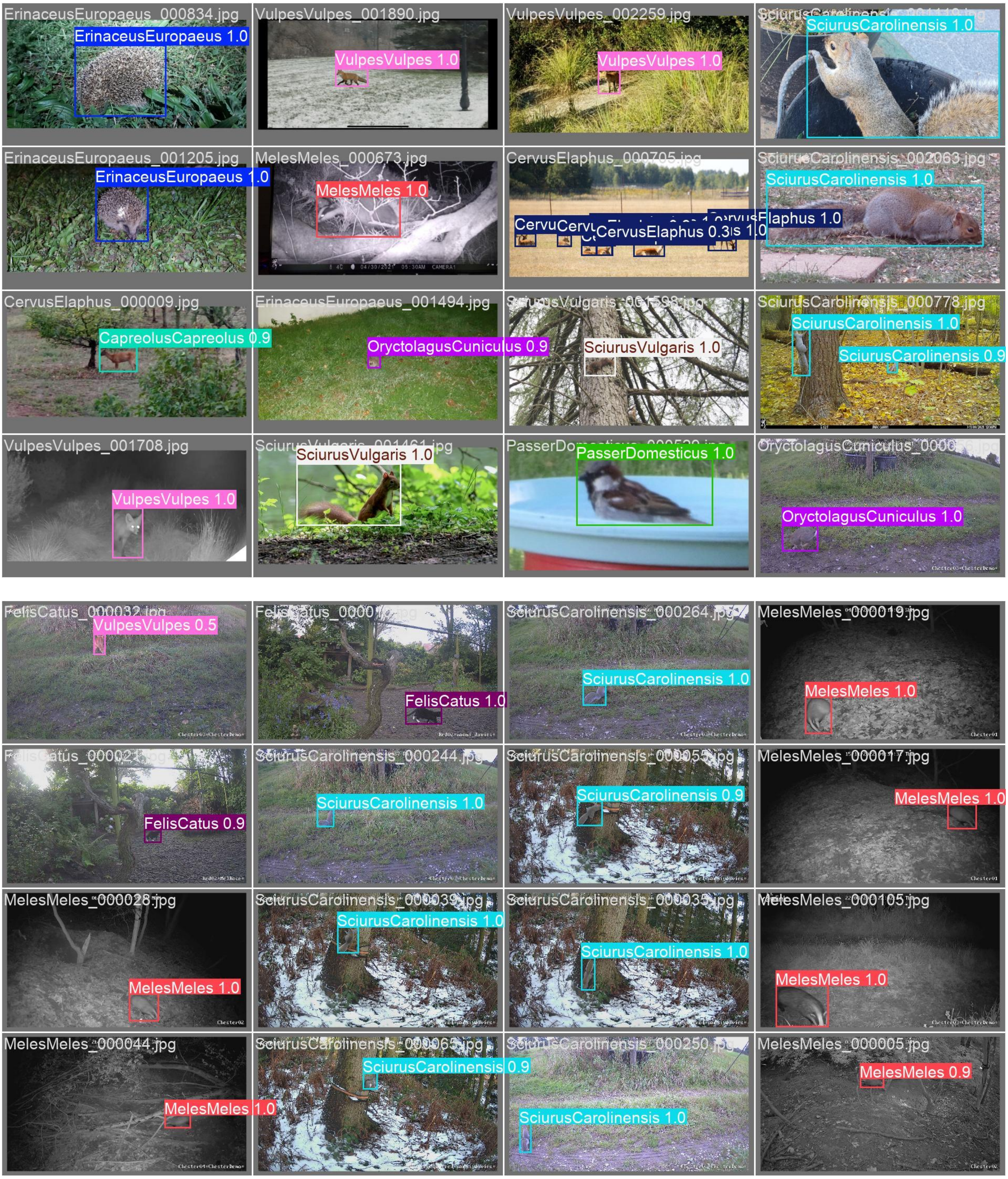

**Appendix B 1:** Validation Batch 0 Prediction

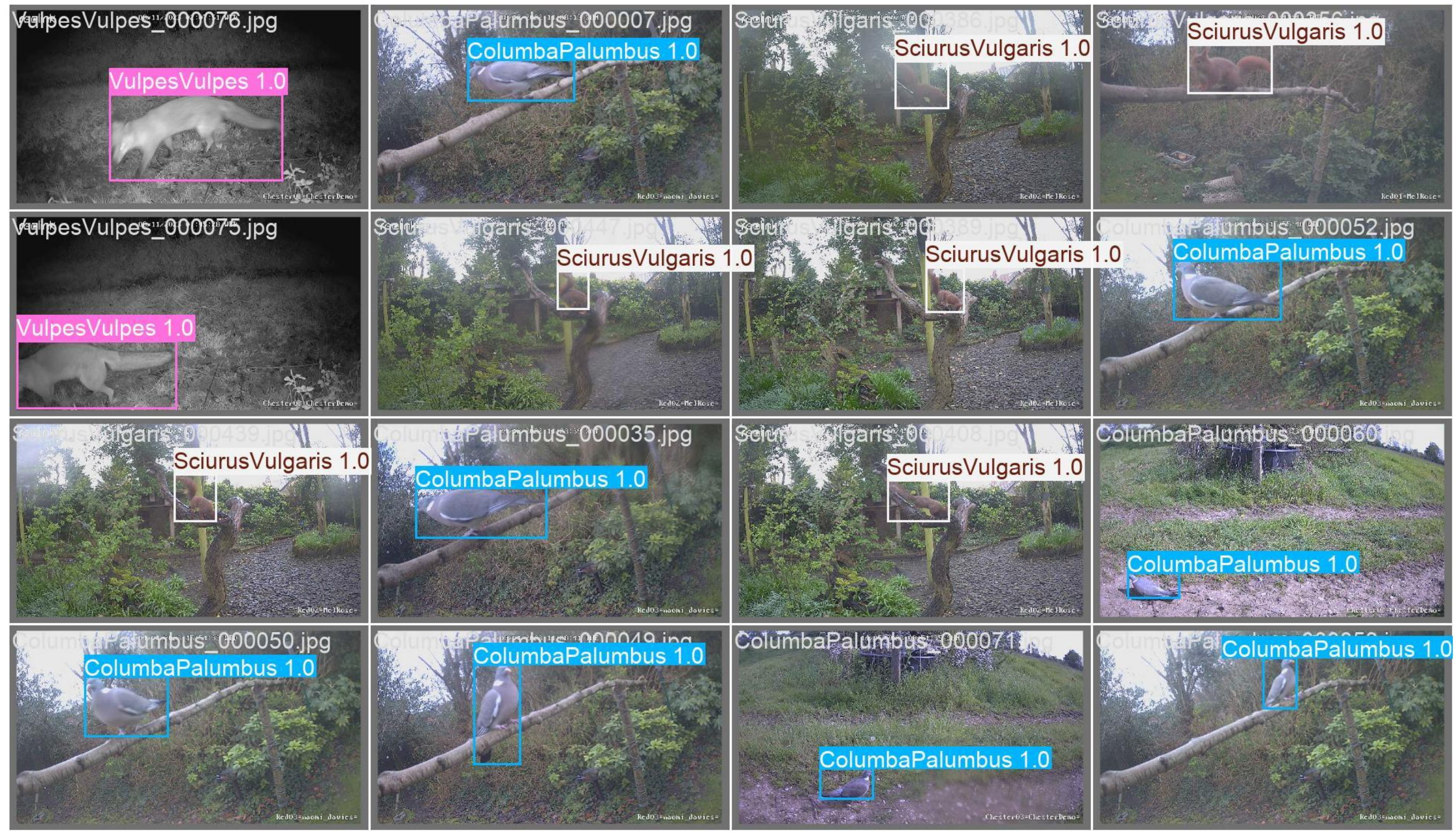

**Appendix B 2:** Validation Batch 1 Prediction

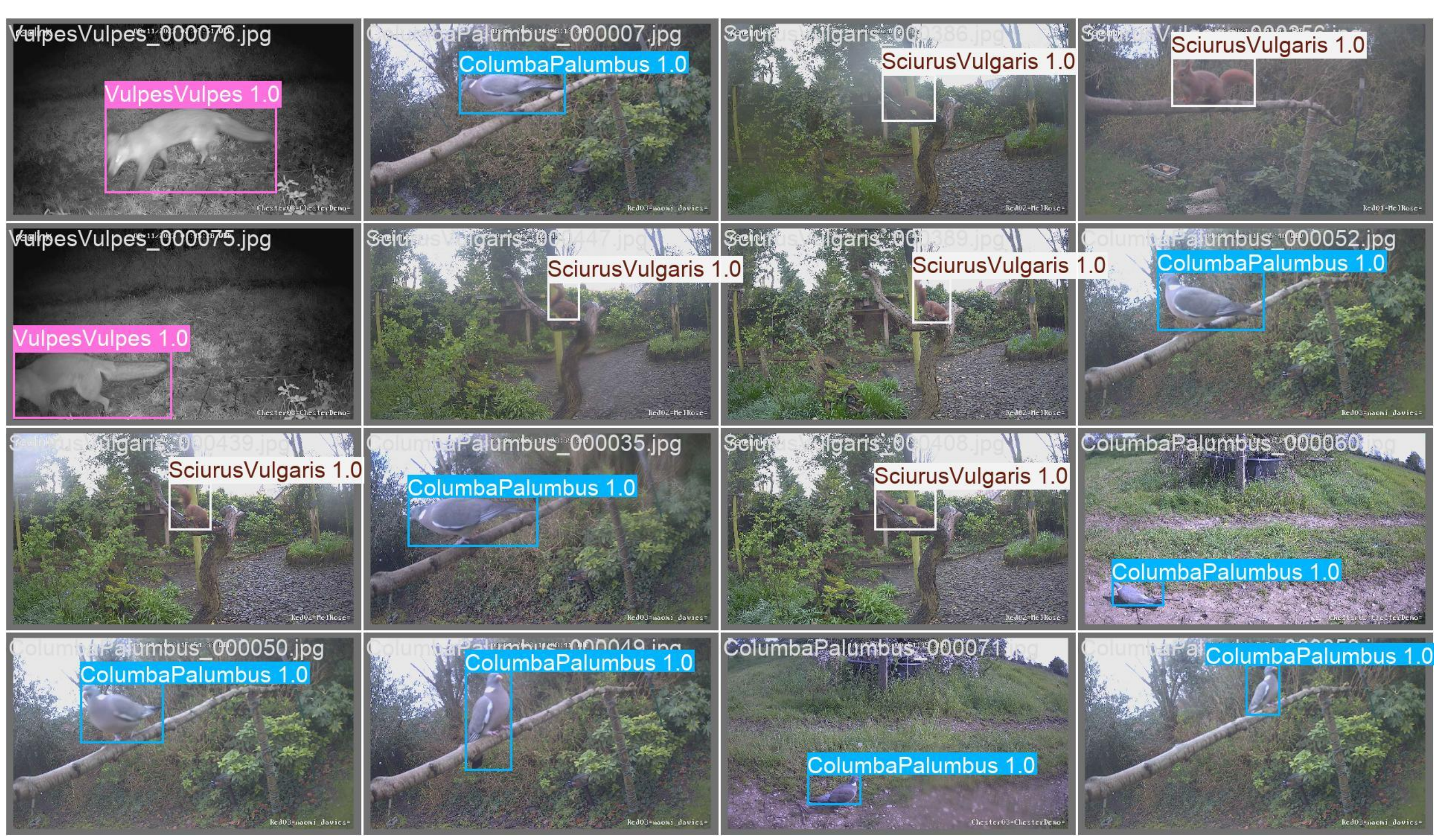

**Appendix B 3:** Validation Batch 2 Prediction

**APPENDIX C:** SAMPLE DETECTIONS FROM THE UK MAMMALS MODEL

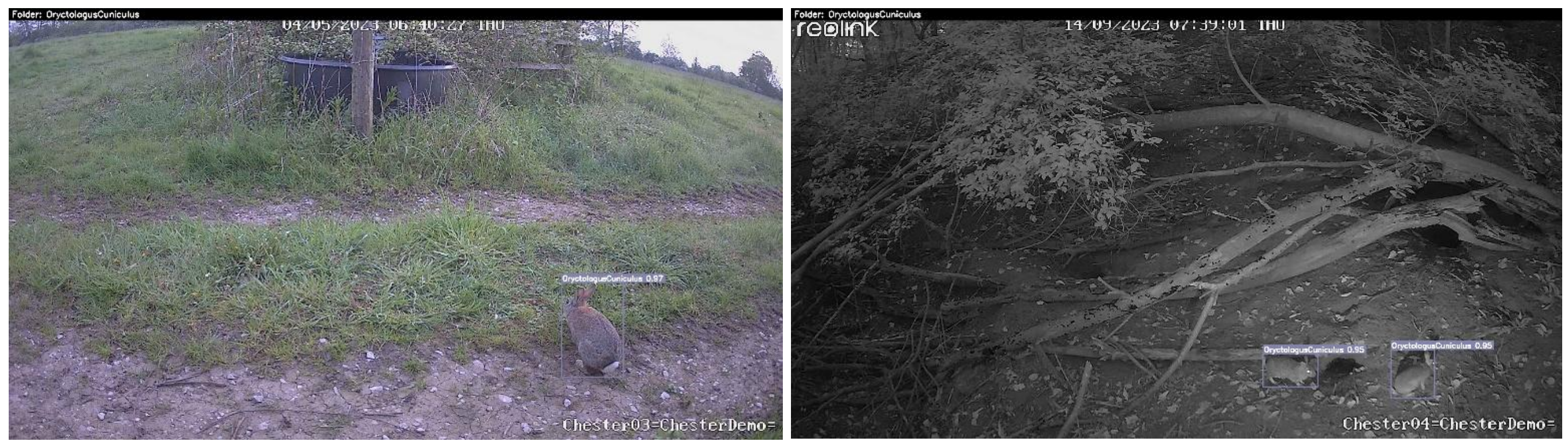

**Appendix C 1:** *Oryctolagus cuniculus*

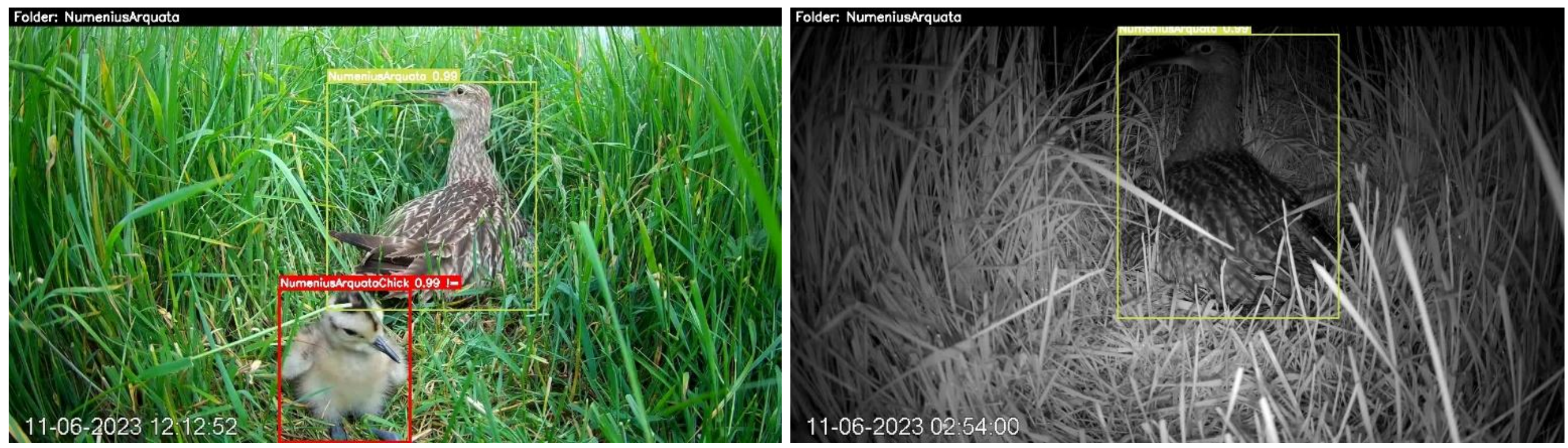

**Appendix C 2:** *Numenius arquata*

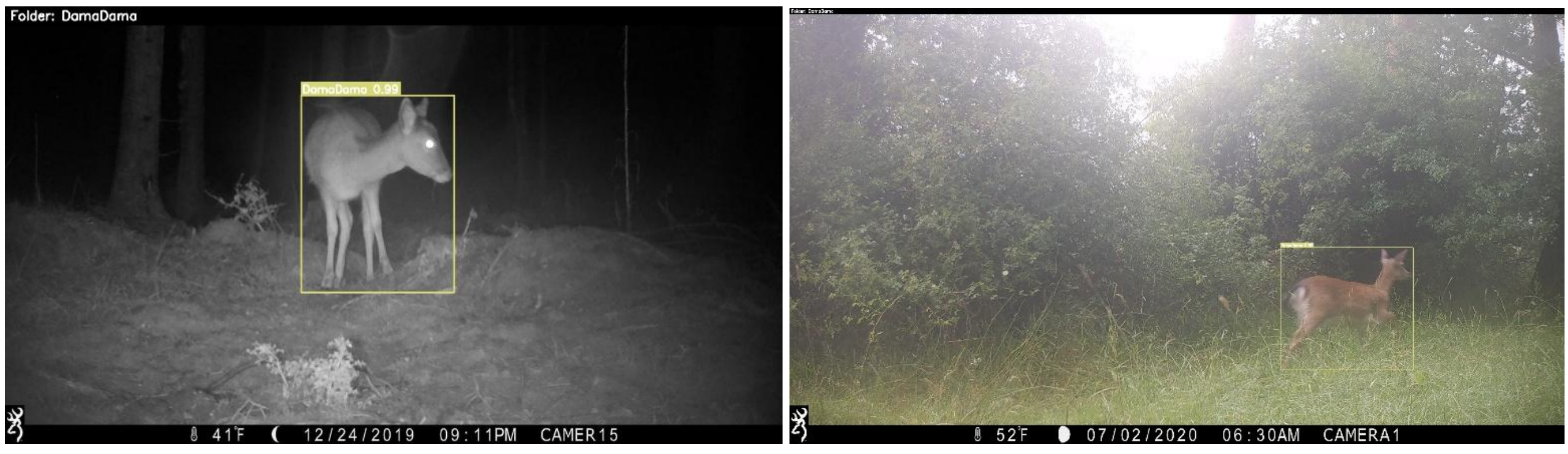

**Appendix C 3:** *Dama dama*

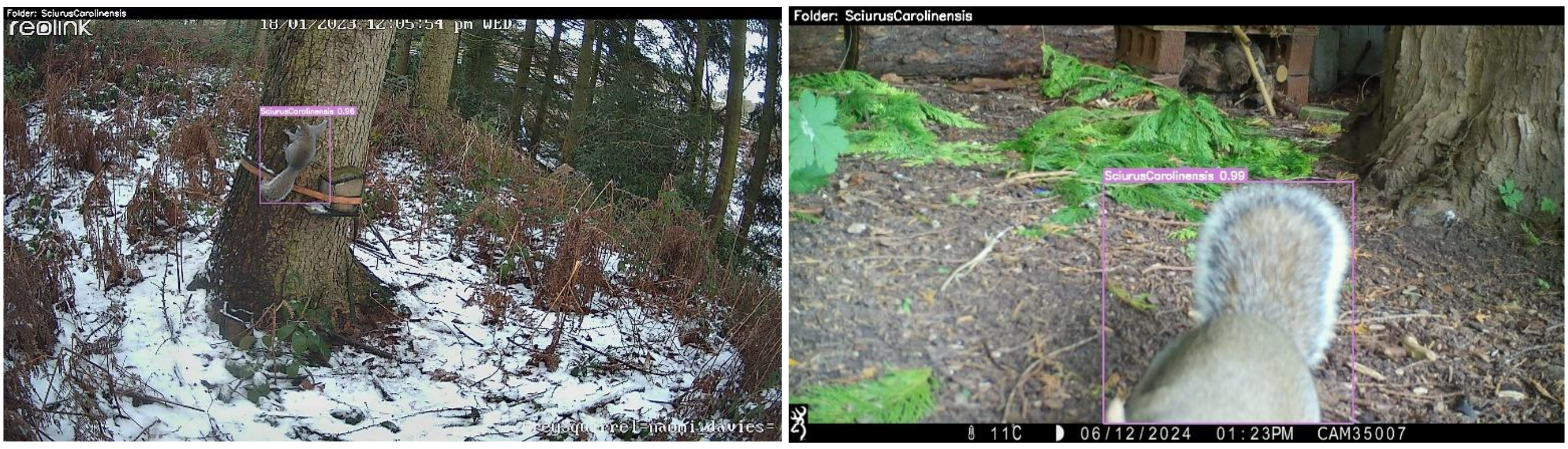

**Appendix C 4:** *Sciurus carolinensis*